\documentclass[autodetect-engine]{article}
\usepackage{tikz}

\usepackage{multicol}
\usepackage[letterpaper,top=2cm,bottom=2cm,left=3cm,right=3cm,marginparwidth=1.75cm]{geometry}

\usepackage{amsmath}
\usepackage{graphicx}
\usepackage{amssymb}
\usepackage{amsfonts}
\usepackage[colorlinks=true, allcolors=blue]{hyperref}
\usepackage{amsthm}
\usepackage{algorithmic}
\usepackage{algorithm}
\usepackage{booktabs}
\usepackage{listings}
\usepackage{xcolor}

\theoremstyle{definition}

\newcommand{\beginsupplement}{
  \setcounter{equation}{0}
  \renewcommand{\theequation}{S-\arabic{equation}}
  \setcounter{table}{0}
  \renewcommand{\thetable}{S-\arabic{table}}
  \setcounter{figure}{0}
  \renewcommand{\thefigure}{S-\arabic{figure}}
  \setcounter{algorithm}{0}
  \renewcommand{\thealgorithm}{S-\arabic{algorithm}}
  \setcounter{lstlisting}{0}
  \renewcommand{\thelstlisting}{S-\arabic{lstlisting}}
}

\definecolor{codegreen}{rgb}{0,0.6,0}
\definecolor{codegray}{rgb}{0.5,0.5,0.5}
\definecolor{codepurple}{rgb}{0.58,0,0.82}
\definecolor{backcolour}{rgb}{0.95,0.95,0.92}

\lstdefinestyle{mystyle}{
    backgroundcolor=\color{backcolour},   
    commentstyle=\color{codegreen},
    keywordstyle=\color{magenta},
    numberstyle=\tiny\color{codegray},
    stringstyle=\color{codepurple},
    basicstyle=\ttfamily\footnotesize,
    breakatwhitespace=false,         
    breaklines=true,                 
    captionpos=b,                    
    keepspaces=true,                 
    numbers=left,                    
    numbersep=5pt,                  
    showspaces=false,                
    showstringspaces=false,
    showtabs=false,                  
    tabsize=2
}

\title{AIJack: Let's Hijack AI! \\ Security and Privacy Risk Simulator for Machine Learning}
\author{Hideaki Takahashi (Koukyosyumei)\\
  \texttt{koukyosyumei@hotmaial.com}}
\date{}

\begin{document}
\maketitle

\begin{abstract}
This paper introduces \textbf{AIJack}, an open-source library designed to assess security and privacy risks associated with the training and deployment of machine learning models. Amid the growing interest in big data and AI, machine learning research and business advancements are accelerating. However, recent studies reveal potential threats, such as the theft of training data and the manipulation of models by malicious attackers. Therefore, a comprehensive understanding of machine learning's security and privacy vulnerabilities is crucial for the safe integration of machine learning into real-world products. AIJack aims to address this need by providing a library with various attack and defense methods through a unified API. The library is publicly available on GitHub (\url{https://github.com/Koukyosyumei/AIJack}).
\end{abstract}



\section{Introduction}


Machine learning (ML) has become a foundational component of diverse applications, spanning image recognition to natural language processing. As these technologies proliferate, comprehending and addressing security risks associated with ML models becomes imperative.

While ML models continuously enhance accuracy, attackers can significantly diminish it by introducing artificially created data. Evasion Attacks~\cite{biggio2013evasion,goodfellow2014explaining}, using specialized algorithms, can induce model malfunctions. Consider a road sign classification model; adding imperceptible noise to a "speed limit 30km" sign may misclassify it as "speed limit 60km." Those perturbed samples are often called adversarial examples. Another attack, Poisoning Attack~\cite{biggio2012poisoning}, injects contaminated data during training, lowering model accuracy.

Various strategies have been proposed to counter such attacks. Certified robustness~\cite{lecuyer2019certified,chen2021cost}, a technique to formally guarantee that adversarial examples cannot lead to undesirable predictions, has gained attention. Another approach is debugging machine learning models~\cite{Kang2018ModelAF,wu2020complaint,10.1145/3132747.3132785}, which aims to identify inputs causing unexpected behaviors.

ML also poses privacy challenges. Collecting large amounts of data for training can infringe on privacy, leading to data breaches. Model Inversion Attacks~\cite{10.1145/2810103.2813677,zhu2019deep} reconstruct training data from pre-trained models, threatening sensitive information. Similar attacks, like Membership Inference Attacks~\cite{shokri2017membership}, aim to determine if a data point is part of the model's training data.

Privacy protection techniques, including differential privacy~\cite{abadi2016deep}, k-anonymity~\cite{1617393}, homomorphic encryption~\cite{paillier1999public}, and distributed learning~\cite{mcmahan2017communication,vepakomma2018split}, have been proposed. Differential privacy prevents individual data inference, while homomorphic encryption enables arithmetic operations on encrypted data. Distributed methods like federated learning (FL)~\cite{mcmahan2017communication} facilitate collaborative learning among data owners without violating data privacy.

Thus, assessing ML models' security and privacy risks and evaluating countermeasure effectiveness is crucial. To simplify such simulations, we propose the open-source software, \textbf{AIJack}, offering various attack and defense methods. AIJack enables the experimentation of numerous combinations of attacks and defenses with simple code. Built on PyTorch~\cite{paszke2019pytorch} and scikit-learn~\cite{pedregosa2011scikit}, users can easily incorporate AIJack into existing code.

\section{Package Design}

\begin{table}[!ht]
\centering
\begin{tabular}{ccc}
\textbf{Type} & \textbf{Subcategory} & \textbf{Method} \\ \hline
FL & Horizontal & \begin{tabular}{c} FedAVG~\cite{mcmahan2017communication}, FedProx~\cite{li2020federated}, FedMD~\cite{li2019fedmd}, \\ FedGEMS~\cite{cheng2021fedgems}, DSFL~\cite{itahara2021distillation}, MOON~\cite{li2021model}, FedEXP~\cite{jhunjhunwala2023fedexp} \end{tabular} \\ \hline
FL & Vertical & SplitNN~\cite{vepakomma2018split}, SecureBoost~\cite{cheng2021secureboost} \\ \hline
Attack & Model Inversion & \begin{tabular}{c} MI-FACE~\cite{10.1145/2810103.2813677}, DLG~\cite{zhu2019deep}, iDLG~\cite{zhao2020idlg}, GS~\cite{geiping2020inverting}, \\ CPL~\cite{wei2020framework}, GradInversion~\cite{yin2021see}, GAN Attack~\cite{hitaj2017deep} \end{tabular} \\ \hline
Attack & Label Leakage & Norm Attack~\cite{li2021label} \\ \hline
Attack & Poisoning & \begin{tabular}{c} History Attack~\cite{cao2022mpaf}, Label Flip~\cite{cao2022mpaf}, \\ MAPF~\cite{cao2022mpaf}, SVM Poisoning~\cite{biggio2012poisoning} \end{tabular} \\ \hline
Attack & Backdoor & DBA~\cite{Xie2020DBADB}, Model Replacement~\cite{pmlr-v108-bagdasaryan20a} \\ \hline
Attack & Free-Rider & Delta-Weight~\cite{lin2019free} \\ \hline
Attack & Evasion & Gradient-Descent Attack~\cite{biggio2013evasion}, FGSM~\cite{goodfellow2014explaining}, DIVA~\cite{hao2022tale} \\ \hline
Attack & Membership Inference & Shadow Attack~\cite{shokri2017membership} \\ \hline
Defense & Homomorphic Encryption & Paillier~\cite{paillier1999public} \\ \hline
Defense & Differential Privacy & DPSGD~\cite{abadi2016deep}, AdaDPS~\cite{li2022private}, DPlis~\cite{wang2021dplis} \\ \hline
Defense & Anonymization & Mondrian~\cite{1617393} \\ \hline
Defense & Robustness & PixelDP~\cite{lecuyer2019certified}, Cost-Aware Robust Tree Ensemble~\cite{chen2021cost} \\ \hline
Defense & Debugging & Model Assertions~\cite{Kang2018ModelAF}, Rain~\cite{wu2020complaint}, Neuron Coverage~\cite{10.1145/3132747.3132785} \\ \hline
Defense & Others & Soteria~\cite{sun2020provable}, FoolsGold~\cite{fung2018mitigating}, MID~\cite{wang2021improving}, Sparse Gradient~\cite{aji-heafield-2017-sparse} \\ \hline
\end{tabular}
\caption{List of Supported Algorithms}
\label{tab:supported-algorithms}
\end{table}

AIJack is designed with the following principles:

\begin{itemize}
    \item \textit{All-around abilities for both attack and defense}: AIJack provides a flexible API for over 40 attack and defense algorithms (see Tab.~\ref{tab:supported-algorithms} for the comprehensive lists). Users can experiment with various combinations of these methods.
    \item \textit{PyTorch-friendly design}: AIJack supports many PyTorch models, allowing integration with minimal modifications to the original code.
    \item \textit{Compatibility with scikit-learn}: AIJack supports many scikit-learn models, enabling easy integration of attacks and defenses.
    \item \textit{Fast Implementation with C++ backend}: AIJack employs a C++ backend for components like Differential Privacy and Homomorphic Encryption, enhancing scalability.
    \item \textit{MPI-Backend for Federated Learning}: AIJack supports MPI-backed federated learning for deployment in High-Performance Computing systems.
    \item \textit{Extensible modular APIs}: AIJack comprises simple modular APIs, allowing easy extension with minimal effort.
\end{itemize}

\paragraph{Examples}

For example, users can easily experiment with evasion attacks and mitigation methods. Code~\ref{code:main:fgsm} shows FGSM, one of the most popular evasion attacks against a neural network, while Code~\ref{code:main:pixeldp} demonstrates PixelDP, one of the most popular mitigation methods against evasion attacks providing certified robustness. These codes can be integrated on the top of PyTorch-based machine learning pipelines.

\begin{figure}[!ht]
\centering
\begin{minipage}{.56\textwidth}
\centering
\begin{lstlisting}[language=Python, caption=FGSM attack against a neural network, label=code:main:fgsm]
from aijack.attack import FGSMAttacker



attacker = FGSMAttacker(nn, criterion, eps=0.3)
perturbed_x = attacker.attack((x, y))
\end{lstlisting}
\end{minipage}\hfill
\begin{minipage}{.41\textwidth}
\centering
\begin{lstlisting}[language=Python, caption=PixelDP, label=code:main:pixeldp]
from aijack.defense import PixelDP

pdp = PixelDP(NN(), num_classes)
pdp.eval()
pred_proba, counts = pdp(data)
pdp.certify(counts)

\end{lstlisting}
\end{minipage}
\end{figure}

Code.~\ref{code:main:miface} shows the example code implementing the MIFACE attack, one of the most popular model inversion attack methods based on gradient descent, allowing attackers to steal training data from the trained model.

\begin{figure}[!ht]
\begin{lstlisting}[language=Python, caption=MIFACE against NN, label=code:main:miface]
from aijack.attack import MI_FACE

attacker = MI_FACE(nn, input_shape)
recovered_data, log = attacker.attack()
\end{lstlisting}
\end{figure}

One possible defense method against such model inversion attacks is training a neural network using Differential Private Stochastic Gradient Descent (DPSGD). DPSGD is easily implemented with AIJack on the top of PyTorch, as illustrated in Code~\ref{code:main:dp}.

\begin{figure}[!ht]
\begin{lstlisting}[language=Python, caption=Deep Learning with Differential Privacy, label=code:main:dp]
from aijack.defense import GeneralMomentAccountant, DPSGDManager

accountant = GeneralMomentAccountant(noise_type="Gaussian", backend="cpp")
privacy_manager = DPSGDManager(accountant, optim.SGD)

accountant.reset_step_info()
dpoptimizer, lot_loader, batch_loader = privacy_manager.privatize(noise_multiplier=sigma)
optimizer = dpoptimizer(nn.parameters(), lr=lr)

for epoch in range(num_iterations):
    for X_lot, y_lot in lot_loader(optimizer):
        for X_batch, y_batch in batch_loader(TensorDataset(X_lot, y_lot)):
            optimizer.zero_grad()
            loss = criterion(nn(X_batch), y_batch)
            loss.backward()
            optimizer.step()
\end{lstlisting}
\end{figure}

AIjack also makes it easy to implement various types of federated learning schemes and simulate their security and privacy risks. Code~\ref{code:main:fedavg} shows the implementation of FedAVG, one of the most standard federated learning protocols, on top of PyTorch. Code~\ref{code:main:mpi} mitigates Code~\ref{code:main:fedavg} to MPI-backend. 

\begin{figure}[!ht]
\centering
\begin{minipage}{.45\textwidth}
\centering
\begin{lstlisting}[language=Python, caption=Standard FedAVG, label=code:main:fedavg]
from aijack.collaborative import (
  FedAVGClient,
  FedAVGServer,
  FedAVGAPI
)











clients = [FedAVGClient(NN()),
           FedAVGClient(NN())]
server = FedAVGServer(clients, NN())

api = FedAVGAPI(
    server, clients,
    criterion, optimizers, dataloaders,
)
api.run()
\end{lstlisting}
\end{minipage}\hfill
\begin{minipage}{.50\textwidth}
\centering
\begin{lstlisting}[language=Python, caption=FedAVG with MPI backend, label=code:main:mpi]
from aijack.collaborative import (
  FedAVGClient, FedAVGServer, MPIFedAVGAPI, 
  MPIFedAVGClientManager, 
  MPIFedAVGServerManager
)

mcm = MPIFedAVGClientManager()
msm = MPIFedAVGServerManager()
MPIClient = mcm.attach(FedAVGClient)
MPIGServer = msm.attach(FedAVGServer)

if myid == 0:
    ids = list(range(1, size))
    server = MPIServer(com, ids, NN())
    api = MPIFedAVGAPI(
        comm, server, True, criterion
    )
else:
    client = MPIClient(comm, NN(), myid)
    api = MPIFedAVGAPI(
        comm, client, False, criterion,
        optimizer, dataloader
    )
    
api.run()
\end{lstlisting}
\end{minipage}
\end{figure}

Then, users can easily incorporate various attack and defense strategies into federated learning by \textit{attaching} those abilities to servers and clients. For instance, Code~\ref{code:main:gradinv} demonstrates a gradient-based model inversion attack against federated learning, while Code~\ref{code:main:paillier} applies Paillier Encryption, one of the most popular Homomorphic Encryption methods, to mitigate such inversion attacks.

\begin{figure}[!ht]
\begin{minipage}{.48\textwidth}
\centering
\begin{lstlisting}[language=Python, caption=Gradient Inversion Attack against FL, label=code:main:gradinv]
from aijack.collaborative import (
  FedAVGClient,
  FedAVGServer,
  FedAVGAPI
)
from aijack.attack import ( 
  GradientInversionAttackServerManager
)





mg = GradientInversionAttackServerManager(
    input_shape,
)
MaliciousServer = mg.attach(FedAVGServer)

clients = [FedAVGClient(NN()),
           FedAVGClient(NN())]
server = MaliciousServer(clients, NN())

api = FedAVGAPI(
    server, clients,
    criterion, optimizers, dataloaders,
)
api.run()
\end{lstlisting}
\end{minipage}\hfill
\begin{minipage}{.47\textwidth}
\centering
\begin{lstlisting}[language=Python, caption=FL with Paillier Encryption, label=code:main:paillier]
from aijack.collaborative import (
  FedAVGClient,
  FedAVGServer,
  FedAVGAPI
)
from aijack.defense import ( 
  PaillierGradientClientManager,
  PaillierKeyGenerator
)

keygenerator = PaillierKeyGenerator(512)
pk, sk = keygenerator.generate_keypair()

mg = PaillierGradientClientManager(pk, sk)
PaillierClient = mg.attach(FedAVGClient)

clients = [PaillierClient(NN()),
           PaillierClient(NN())]
server = FedAVGServer(clients, NN())

api = FedAVGAPI(
    server,
    clients,
    criterion,
    optimizers,
    dataloaders,
)
api.run()
\end{lstlisting}
\end{minipage}
\end{figure}

Additional examples are available in the Appendix and our documentation\footnote{\url{https://koukyosyumei.github.io/AIJack/}}.

\paragraph{Comparison with other libraries}

While many existing libraries prioritize security and privacy within the realm of machine learning, AIJack distinguishes itself by offering unparalleled flexibility and a wide array of options for exploring security and privacy attacks and their corresponding countermeasures. Unlike many prior tools that focus on specific types of attacks and defenses—such as ART~\cite{nicolae2018adversarial}, foolbox~\cite{rauber2017foolbox}, AdvBox~\cite{goodman2020advbox}, and advertorch~\cite{ding2019advertorch} for adversarial examples; ml\_privacy\_meter~\cite{murakonda2020ml} and PrivacyRaven~\cite{privacyraven} for model inversion and membership inference attacks; tensorflow-privacy~\cite{tensorflowprivacy} and Opacus~\cite{yousefpour2021opacus} for differential privacy; and FATE~\cite{fate}, flower~\cite{beutel2020flower}, FedML~\cite{he2020fedml}, VFLAIR~\cite{zou2023vflair}, and secretflow~\cite{secretflow} for federated learning—AIJack stands out by encompassing all of these and more. This comprehensive approach empowers users to simulate various attack scenarios easily and experiment with multiple defense methods simultaneously.

\section*{Acknowledgement}

We acknowledge every contributor for their support of this project. The comprehensive list of contributors is in our GitHub repository.

{\small
\bibliographystyle{unsrt}
\bibliography{ref}
}

\beginsupplement
\appendix

\section{Standard Machine Learning}

This section explores the vulnerabilities and defenses of standard machine learning. We delve into various attack techniques, including adversarial examples, poisoning attacks, model inversion attacks, and membership inference attacks. We then discuss defense mechanisms like certified robustness, debugging methods, differential privacy, and k-anonymity, all crucial for building secure and trustworthy machine learning systems.

Furthermore, practical examples and code snippets are provided to illustrate the implementation of both attack and defense strategies with AIJack. Since some detailed arguments and parameters are committed for better readability, it might be better for interested readers to check the detailed documentation at \url{https://koukyosyumei.github.io/AIJack/}.

\subsection{Attack}

\paragraph{Adversarial Examples}

Adversarial examples are carefully crafted inputs aimed at deceiving machine learning models~\cite{zhang2019adversarial}. These subtle alterations, often undetectable to humans, can lead the model to produce noticeably incorrect outputs. This form of attack is also known as an evasion attack. For example, an adversarial example could be a slightly altered image that causes a computer vision system to misidentify it as an entirely different object. Consequently, adversarial examples raise concerns about the reliability and applicability of machine learning models in real-world scenarios. Ongoing research in this domain focuses on techniques for generating and defending against adversarial examples, promoting the development of more resilient and dependable machine learning systems.

Code~\ref{code:evasion} shows the example codes implementing the gradient-based evasion attack~\cite{biggio2013evasion} against SVM trained with scikit-learn, and Code~\ref{code:fgsm} demonstrates the FGSM attack~\cite{goodfellow2014explaining} against a neural network implemented with PyTorch.

\begin{figure}[!ht]
\centering
\begin{minipage}{.48\textwidth}
\centering
\begin{lstlisting}[language=Python, caption=Evasion Attack against SVM, label=code:evasion]
from aijack.attack import (
    Evasion_attack_sklearn
)

attacker = Evasion_attack_sklearn(
    clf, initial_data_positive
)
malicious_data, log = attacker.attack(
    initial_data_negative
)
\end{lstlisting}
\end{minipage}\hfill
\begin{minipage}{.48\textwidth}
\centering
\begin{lstlisting}[language=Python, caption=FGSM attack against NN, label=code:fgsm]
from aijack.attack.evasion import FGSMAttacker



attacker = FGSMAttacker(
    nn, criterion, eps=0.3
)
perturbed_x = attacker.attack((x_origin, y_origin))
\end{lstlisting}
\end{minipage}
\end{figure}

\paragraph{Poisoning Attack}

In machine learning, poisoning attacks are a significant threat aimed at manipulating a model's training process toward malicious ends~\cite{tian2022comprehensive}. They often involve injecting corrupted data points, targeting specific classes to reduce accuracy for certain predictions, or even altering the model's decision boundaries. For example, injecting mislabeled images into a facial recognition model could lead to the misidentification of specific individuals.

Code~\ref{code:poisoning} is the example code of the Poisoning Attack against SVM~\cite{biggio2012poisoning} trained with scikit-learn.

\begin{figure}[!ht]
\begin{lstlisting}[language=Python, caption=Poisoning Attack against SVM, label=code:poisoning]
from aijack.attack import Poison_attack_sklearn

attacker = Poison_attack_sklearn(clf, X_train, y_train)
malicious_data, log = attacker.attack(initial_data, 1, X_valid, y_valid)
\end{lstlisting}
\end{figure}

\paragraph{Model Inversion Attack}

Model inversion attacks present a serious security risk in machine learning, especially with sensitive data. They involve reconstructing training data by analyzing a model's parameters and outputs, often using techniques like gradient descent on carefully crafted queries~\cite{song2022survey}. Successful attacks can compromise sensitive information in the training data, risking user privacy or disclosing confidential details.

Code.~\ref{code:miface} shows the example code implementing the MIFACE attack~\cite{10.1145/2810103.2813677}, one of the most popular model inversion attack methods based on gradient descent.

\begin{figure}[!ht]
\begin{lstlisting}[language=Python, caption=MIFACE against NN, label=code:miface]
from aijack.attack import MI_FACE

mi = MI_FACE(nn, input_shape)
x_result_1, log = mi.attack()
\end{lstlisting}
\end{figure}

\paragraph{Membership Inference Attack}

Machine learning models are also vulnerable to membership inference attacks (MIAs), where attackers try to ascertain if a data point is part of the model's training dataset~\cite{hu2022membership}. By analyzing the model's outputs, like predictions or confidence scores, attackers can craft queries to discern membership. Deviations in the model's behavior for a specific data point compared to unseen data may reveal membership in the training set. MIAs are a significant privacy concern, particularly with sensitive datasets such as medical records or financial transactions.

Code.~\ref{code:shadow} shows the example implementation of a shadow attack~\cite{shokri2017membership}, which is one of the earliest and most popular membership inference attacks.

\begin{figure}[!ht]
\begin{lstlisting}[language=Python, caption=Membership Inference Attack, label=code:shadow]
from aijack.attack.membership import ShadowMembershipInferenceAttack

shadow_models = [clf() for _ in range(2)]
attack_models = [clf() for _ in range(10)]

attacker = ShadowMembershipInferenceAttack(clf, shadow_models, attack_models)
attacker.fit(X_shadow, y_shadow)
\end{lstlisting}
\end{figure}

\subsection{Defense}

\paragraph{Certified Robustness}

Certified robustness measures a model's assured resilience against adversarial examples, offering formal guarantees beyond traditional accuracy metrics~\cite{silva2020opportunities,li2023sok}. To achieve this, methods like randomized smoothing inject noise into the internal process, though it may lower accuracy and computational efficiency. Ongoing research aims to enhance certification methods, balancing robustness, accuracy, and efficiency for practical machine-learning applications.

Code~\ref{code:pixeldp} is the example of PixelDP~\cite{lecuyer2019certified} applied for a neural network implemented with PyTorch. Code~\ref{code:cost-tree} also shows Cost-Aware Robust Tree Ensemble~\cite{chen2021cost}, which is the tree-based certified robust training method.

\begin{figure}[!ht]
\centering
\begin{minipage}{.38\textwidth}
\centering
\begin{lstlisting}[language=Python, caption=PixelDP, label=code:pixeldp]
from aijack.defense import (
    PixelDP
)

pdp = PixelDP(
    nn,
    num_classes,
    eps=0.3,
    delta=1e-5,
)

# Normal Training of PyTorch

pdp.eval()
pred_proba, counts = pdp(data)
pdp.certify(counts)

\end{lstlisting}
\end{minipage}\hfill
\begin{minipage}{.57\textwidth}
\centering
\begin{lstlisting}[language=Python, caption=Cost-Aware Robust Tree Ensemble, label=code:cost-tree]
from aijack.collaborative.tree import (
    XGBoostClassifierAPI,
    XGBoostClient,
)



p0 = XGBoostClient(X_train)

# Set the attack-cost constraint to each feature
p0.set_cost_constraint_map(constraints)

clf = XGBoostClassifierAPI()
clf.fit([p0], y_train.tolist())

pred_proba = clf.predict_proba(X_test)
\end{lstlisting}
\end{minipage}
\end{figure}

\paragraph{Debugging}

Debugging neural networks is uniquely challenging due to their complex, non-linear nature. Neuron coverage~\cite{10.1145/3132747.3132785}, akin to code coverage in software engineering, measures the proportion of activated neurons above a predefined threshold during testing. Higher coverage reflects a more thorough exploration of the network's decision space. This method can be utilized to detect unknown adversarial examples and unexpected behaviors of a neural network caused by poisoning attacks. 

With AIJack, users can easily implement Neuron Coverage, as shown in Code~\ref{code:nc}.

\begin{figure}[!ht]
\begin{lstlisting}[language=Python, caption=Debugging of a neural network with Neuron Coverage, label=code:nc]
from aijack.defense.debugging.neuroncoverage import *

NCT = neuroncoverage.NeuronCoverageTracker(nn, threshold, dummy_data)
nc = NCT.coverage(debug_dataloader)
\end{lstlisting}
\end{figure}

\paragraph{Differential Privacy}

Differential privacy (DP) offers a formal privacy guarantee in randomized algorithms~\cite{dwork2008differential}. It ensures that the inclusion or exclusion of a single data point has minimal impact on the algorithm's output. Applying differential privacy to machine learning protects individual privacy while still enabling the training of effective models~\cite{el2022differential}. This method can be the solution against the model inversion attack and membership inference attacks.

Differentially, Private Stochastic Gradient Descent (DPSGD)~\cite{abadi2016deep} is a prominent technique that injects noise into the gradient updates during model training. This added noise safeguards individual data points while introducing a controllable privacy-utility trade-off. DPSGD serves as a cornerstone for privacy-preserving machine learning, allowing researchers to train models on sensitive data without compromising individual privacy.

Code~\ref{code:dpsgd} illustrates the standard workload of training a neural network using Differential Private Stochastic Gradient Descent (DPSGD). AIJack also supports several variants of DPSGD, including AdaDPS~\cite{li2022private} and DPlis~\cite{wang2021dplis}. Our API's design draws inspiration from Opacus~\cite{yousefpour2021opacus}, a renowned framework dedicated to deep learning with differential privacy. Additionally, our AIJack facilitates a more intuitive implementation, closely aligning with the algorithm initially proposed in~\cite{abadi2016deep}.

\begin{figure}[!ht]
\begin{lstlisting}[language=Python, caption=Deep Learning with Differential Privacy, label=code:dpsgd]
from aijack.defense import GeneralMomentAccountant, DPSGDManager

accountant = GeneralMomentAccountant(noise_type="Gaussian", backend="cpp")
privacy_manager = DPSGDManager(accountant, optim.SGD)
# privacy_manager = AdaDPSManager(accountant, optim.SGD)

accountant.reset_step_info()
dpoptimizer_cls, lot_loader, batch_loader = privacy_manager.privatize(
    noise_multiplier=sigma
)
optimizer = dpoptimizer_cls(nn.parameters(), lr=lr)

for epoch in range(num_iterations):
    for X_lot, y_lot in lot_loader(optimizer):
        for X_batch, y_batch in batch_loader(TensorDataset(X_lot, y_lot)):
            optimizer.zero_grad()
            loss = criterion(nn(X_batch), y_batch)
            loss.backward()
            optimizer.step()
\end{lstlisting}
\end{figure}

\paragraph{K-Anonymity}

K-anonymity is a data anonymization method that aims to safeguard individual privacy within datasets~\cite{1617393}. It ensures each record is indistinguishable from at least k-1 others based on identifying attributes, such as zip code or age range. While offering a 'hiding in the crowd' privacy safeguard, achieving k-anonymity may require data modifications like suppression or generalization, which can impact dataset usability. Balancing privacy and data utility remains a key challenge in k-anonymity research. 

AIJack supports the Mondrian algorithm~\cite{1617393}, one of the most popular k-anonymity methods. Code~\ref{code:kano} provides a simple example of the implementation of Mondrian.

\begin{figure}[!ht]
\begin{lstlisting}[language=Python, caption=K-Anonymity, label=code:kano]
from aijack.defense.kanonymity import Mondrian

mondrian = Mondrian(k=2)
adf_ignore_unused_features = mondrian.anonymize(df, feature_columns, sensitive_column, is_continuous_map)
\end{lstlisting}
\end{figure}

\section{Federated Learning}

Federated learning (FL) revolutionizes machine learning by facilitating collaborative model training on distributed devices while preserving data privacy~\cite{mammen2021federated}. Unlike centralized learning, where raw data is sent to a central server, FL enables devices like smartphones or sensors to train models locally on their data. Only model updates, not raw data, are shared with a central server for aggregation, ensuring privacy. This iterative refinement process enhances the global model without compromising individual data privacy. There are two main approaches: horizontal and vertical.

This section delves into the security vulnerabilities and defense mechanisms within federated learning. We explore various attack techniques, including inversion, label leakage, poisoning, and free-rider attacks. We then discuss defense mechanisms like homomorphic encryption, model distillation, gradient modification, and abnormal client detection, all crucial for building secure and robust federated learning systems.

\paragraph{Horizontal Federated Learning}

Horizontal federated learning works best when all devices have the same type of data, like different users' smartphone photos. Each device trains the model on its own photos but only shares the model updates, not the actual photos.

The most naive and popular horizontal federated learning protocol is FedAvg~\cite{mcmahan2017communication}, which utilizes the average of local updates as the aggregation method. AIJack also supports FedProx~\cite{li2020federated}, which introduces a proximal term to mitigate the impact of data heterogeneity across devices, a common challenge in FL. In addition, AIJack implements various popular protocols like MOON~\cite{li2021model} and FedEXP~\cite{jhunjhunwala2023fedexp}. Those protocols are easily implemented and combined with AIJack, as shown in Code~\ref{code:hfl}. AIJack also supports MPI as the backend for parallel computing so that users can run massive experiments in high-performance computing clusters (see Code~\ref{code:mpi:hfl}).

\begin{figure}[!ht]
\begin{lstlisting}[language=Python, caption=Horizontal Federated Learning, label=code:hfl]
from aijack.collaborative import FedAVGClient, FedAVGServer, FedAVGAPI



clients = [FedAVGClient(NN()), FedAVGClient(NN())]
# clients = [FedKDClient(NN()), FedKDClient(NN())]
# clients = [MOONClient(NN()), MOONClient(NN())]
# clients = [FedProxClient(NN()), FedProxClient(NN())]
           
server = FedAVGServer(clients, NN())
# server = FedEXPServer(clients, NN())

api = FedAVGAPI(server, clients, criterion, optimizers, dataloaders,)
# api = FedProxAPI(server, clients, criterion, optimizers, dataloaders)
api.run()
\end{lstlisting}
\end{figure}

\begin{figure}[!ht]
\begin{lstlisting}[language=Python, caption=FL with MPI backend, label=code:mpi:hfl]
from aijack.collaborative import (
    FedAVGClient, FedAVGServer, 
    MPIFedAVGAPI, MPIFedAVGClientManager, MPIFedAVGServerManager
)

mcm = MPIFedAVGClientManager()
msm = MPIFedAVGServerManager()
MPIClient = mcm.attach(FedAVGClient)
MPIGServer = msm.attach(FedAVGServer)

if myid == 0:
    client_ids = list(range(1, size))
    server = MPIServer(comm, client_ids, model)
    api = MPIFedAVGAPI(comm, server, True, criterion)
else:
    client = MPIClient(comm, model, user_id=myid)
    api = MPIFedAVGAPI(comm, client, False, criterion,optimizer, dataloader)
    
api.run()
\end{lstlisting}
\end{figure}

\paragraph{Vertical Federated Learning}

Vertical federated learning tackles situations where different parties hold complementary data~\cite{liu2024vertical}. For instance, a hospital might have patient diagnoses, while a fitness tracker company has activity data for the same individuals. Here, both parties contribute their unique data aspects to collaboratively train a model for personalized health insights without ever revealing the complete individual information.

One of the most popular vertical federated learning protocols is SplitNN (split learning)~\cite{vepakomma2018split}, which assumes that one party has training labels and the other party owns the features. SplitNN achieves privacy-preserving training by splitting a neural network and exchanging the intermediate values during training. AIJack helps users easily implement SplitNN, as shown in Code~\ref{code:splitnn}.

\begin{figure}[!ht]
\begin{lstlisting}[language=Python, caption=SplitNN, label=code:splitnn]
from aijack.collaborative.splitnn import SplitNNAPI, SplitNNClient

manager = NormAttackSplitNNManager(criterion, device="CPU")

client_1 = SplitNNClient(NN_1(), user_id=0)
client_2 = SplitNNClient(NN_2(), user_id=0)
clients = [client_1, client_2]

SplitNNAPI = manager.attach(SplitNNAPI)
splitnn = SplitNNAPI(clients, optimizers, train_loader, criterion, num_communication)
splitnn.run()
\end{lstlisting}
\end{figure}

\subsection{Attack}

\paragraph{Inversion Attack}

Federated learning aims to protect data privacy, but it creates a new risk during training. When models share updated gradients with each other, they leak information through these updates~\cite{zhu2019deep}. In a model inversion attack, the attacker tries to guess the original training data by analyzing the gradients. They typically do this by starting with a random seed input and then iteratively refining it until it produces gradients that closely resemble the ones received from the victim client~\cite{huang2021evaluating}.

AIJack allows the implementation of various gradient-based inversion attacks against Federated Learning, including DLG~\cite{zhu2019deep}, iDLG~\cite{zhao2020idlg}, GS~\cite{geiping2020inverting}, CPL~\cite{wei2020framework} and GradInversion~\cite{yin2021see}, with the unified API, as shown in Code~\ref{code:gradinversion}.

\begin{figure}[!ht]
\centering
\centering
\begin{lstlisting}[language=Python, caption=Gradient Inversion Attack against FL, label=code:gradinversion]
from aijack.collaborative import FedAVGClient, FedAVGServer, FedAVGAPI
from aijack.attack import GradientInversionAttackServerManager

manager = GradientInversionAttackServerManager(input_shape, distancename="l2")
DLGFedAVGServer = manager.attach(FedAVGServer)

# manager = GradientInversionAttackServerManager(input_shape, distancename="cossim")
# GSFedAVGServer = manager.attach(FedAVGServer)

# manager = GradientInversionAttackServerManager(input_shape, distancename="l2",                                                      
#                                                optimize_label=False)
# iDLGFedAVGServer = manager.attach(FedAVGServer)

# manager = GradientInversionAttackServerManager(input_shape, distancename="l2",                                                          
#                                      optimize_label=False, lm_reg_coef=0.01)
# CPLFedAVGServer = manager.attach(FedAVGServer)

clients = [FedAVGClient(NN()), FedAVGClient(NN())]
server = DLGFedAVGServer(clients, NN())

api = FedAVGAPI(server, clients, criterion, optimizers, dataloaders)
api.run()
\end{lstlisting}
\end{figure}

\paragraph{Label Leakage Attack}

While split learning allows the label owner and feature owners to collaboratively train a model without directly sharing their data, it is known that the malicious client might be able to estimate the training labels by analyzing the shared intermediate representations~\cite{li2021label}.

Code~\ref{code:normattack} shows the example code implementing such an attack with AIJack.

\begin{figure}[!ht]
\begin{lstlisting}[language=Python, caption=Label Leakage Attack against Split Learning, label=code:normattack]
from aijack.attack.labelleakage import NormAttackSplitNNManager
from aijack.collaborative.splitnn import SplitNNAPI

manager = NormAttackSplitNNManager(criterion, device="cpu")
NormAttackSplitNNAPI = manager.attach(SplitNNAPI)
normattacksplitnn = NormAttackSplitNNAPI(
    clients, optimizers, train_loader, criterion, num_communication
)

normattacksplitnn.run()
\end{lstlisting}
\end{figure}

\paragraph{Poisoning Attack}

Federated learning relies on distributed updates from participating devices, making it vulnerable to poisoning attacks~\cite{cao2022mpaf}. In these attacks, malicious actors compromise some devices and manipulate their updates to influence the global model. This manipulation can involve injecting irrelevant data or altering gradients during training. For example, an attacker might try to reduce the model's accuracy for a certain class by consistently sending updates that weaken its ability to identify that class.

With AIJack, you can easily simulate poisoning attacks within federated learning by attaching those abilities to clients, as demonstrated in Code~\ref{code:poison-fl}.

\begin{figure}[!ht]
\begin{lstlisting}[language=Python, caption=Poisoning attack against FL, label=code:poison-fl]
from aijack.collaborative import FedAVGClient, FedAVGServer, FedAVGAPI
from aijack.attack.poison import MAPFClientWrapper

manager = MAPFClientWrapper(lam=3)
MAPFFedAVGClient = manager.attach(FedAVGClient)

clients = [FedAVGClient(NN()), MAPFFedAVGClient(NN())]
server = FedAVGServer(clients, NN())

api = FedAVGAPI(server, clients, criterion, optimizers, dataloaders)
api.run()
\end{lstlisting}
\end{figure}

\paragraph{Free-rider Attack}

Numerous studies suggest different incentive methods to encourage more data owners to participate in federated learning~\cite{zeng2021comprehensive}. Yet, these structures are vulnerable to free-rider attacks, where malicious clients seek to join federated learning without providing useful updates~\cite{lin2019free}. Free riders may submit random updates or refrain from training entirely, aiming to obtain rewards and model access without bearing local training costs. These attacks hinder overall convergence and performance by depriving the global model of essential training signals.

Users of AIJack can easily experiment with the free-rider attack by attaching it to normal federated learning pipelines, as shown in Code~\ref{code:freerider}.

\begin{figure}[!ht]
\begin{lstlisting}[language=Python, caption=Free-rider attack against FL, label=code:freerider]
from aijack.collaborative import FedAVGClient, FedAVGServer, FedAVGAPI
from aijack.attack.freerider import FreeRiderClientManager

manager = FreeRiderClientManager(mu=0, sigma=1.0)
FreeRiderFedAVGClient = manager.attach(FedAVGClient)

clients = [FedAVGClient(NN()), FreeRiderFedAVGClient(NN())]
server = FedAVGServer(clients, NN())

api = FedAVGAPI(server, clients, criterion, optimizers, dataloaders)
api.run()
\end{lstlisting}
\end{figure}

\subsection{Defense}

\paragraph{Homomorphic Encryption}

As discussed in the previous section, communicating gradients might lead to severe privacy leakage via gradient inversion attacks. One possible mitigation method is encrypting communicated gradients with homomorphic encryptions, which allow the addition of encrypted data~\cite{fang2021privacy,zhang2020privacy}. This enables the central server in federated learning to perform secure aggregation, significantly enhancing privacy.

AIJack has its own implementation of Paillier Encryption~\cite{paillier1999public}, one of the most popular Homomorphic Encryption methods. Users can easily integrate Paillier Encryption with federated learning, as shown in Code~\ref{code:fl:paillier}.

\begin{figure}[!ht]
\begin{lstlisting}[language=Python, caption=FL with Paillier Encryption, label=code:fl:paillier]
from aijack.collaborative import FedAVGClient, FedAVGServer, FedAVGAPI
from aijack.defense import PaillierGradientClientManager, PaillierKeyGenerator

keygenerator = PaillierKeyGenerator(512)
pk, sk = keygenerator.generate_keypair()

mg = PaillierGradientClientManager(pk, sk)
PaillierClient = mg.attach(FedAVGClient)

clients = [PaillierClient(NN()), PaillierClient(NN())]
server = FedAVGServer(clients, NN())

api = FedAVGAPI(server, clients, criterion, optimizers, dataloaders)
api.run()
\end{lstlisting}
\end{figure}

\paragraph{Model Distillation}

Another approach to address training data leakage from gradients is federated learning with Model Distillation (FedMD)~\cite{li2019fedmd}. With FedMD, instead of sending gradients, each client transmits output logits on the public dataset. Subsequently, the central server trains a global model by utilizing the aggregated output logits from clients as soft labels. This method can be seen as a blend of knowledge distillation and federated learning. FedMD also offers flexibility by permitting each client to train its own model structure, as they are not required to use identical models.

AIJack provides the API of FedMD, similar to APIs of standard federated learning protocols, as demonstrated in Code~\ref{code:fedmd}.

\begin{figure}[!ht]
\begin{lstlisting}[language=Python, caption=Federated Learning with Model Distillation, label=code:fedmd]
from aijack.collaborative.fedmd import FedMDAPI, FedMDClient, FedMDServer

clients = [
    FedMDClient(NN_1(), public_dataloader, output_dim=10, user_id=0),
    FedMDClient(NN_2(), public_dataloader, output_dim=10, user_id=1)
]
server = FedMDServer(clients, NN_3())

api = FedMDAPI(server, clients)
api.run()
\end{lstlisting}
\end{figure}

\paragraph{Gradient Modification}

In addition, each client also can modify the local gradient before sending it to reduce the leakage risk. For example, one of the naive methods is sparse gradient~\cite{aji-heafield-2017-sparse}, where the client shares only significant elements within the gradients with the server to prevent the transmitted gradients from being too informative about the training data. \cite{sun2020provable} proposes a more sophisticated approach called Soteria, which has a theoretical guarantee about the reconstruction error. Soteria tries to prevent data leakage from gradients by calculating gradients on the sparsed intermediate representations.

Those gradient modification strategies can be easily added to federated learning protocols with minor modifications of codes, as shown in Code~\ref{code:sparse} and Code~\ref{code:soteria}.

\begin{figure}[!ht]
\begin{lstlisting}[language=Python, caption=FL with Sparse Gradient, label=code:sparse]
from aijack.collaborative.fedavg import FedAVGAPI, FedAVGClient, FedAVGServer
from aijack.defense.sparse import (
    SparseGradientClientManager,
    SparseGradientServerManager,
)

client_manager = SparseGradientClientManager(k=0.3)
SparseGradientFedAVGClient = client_manager.attach(FedAVGClient)
server_manager = SparseGradientServerManager()
SparseGradientFedAVGServer = server_manager.attach(FedAVGServer)

clients = [
    SparseGradientFedAVGClient(NN(), server_side_update=False)
    for i in range(client_num)
]
server = SparseGradientFedAVGServer(clients, NN(), server_side_update=False)

api = FedAVGAPI(server, clients, criterion, optimizers, dataloaders)
api.run()
\end{lstlisting}
\end{figure}

\begin{figure}[!ht]
\begin{lstlisting}[language=Python, caption=FL with Soteria, label=code:soteria]
from aijack.collaborative.fedavg import FedAVGAPI, FedAVGClient, FedAVGServer
from aijack.defense import SoteriaClientManager

manager = SoteriaClientManager("conv", "lin", target_layer_name="lin.0.weight")
SoteriaFedAVGClient = manager.attach(FedAVGClient)

clients = [SoteriaFedAVGClient(NN()) for i in range(client_num)]
server = FedAVGServer(clients, global_model)

api = FedAVGAPI(server, clients, criterion, optimizers, dataloaders)
api.run()
\end{lstlisting}
\end{figure}

\paragraph{Abnormal Client Detection}

To prevent poisoning and free-rider attacks, the central server must detect abnormal clients that maliciously behave. Techniques like FoolsGold~\cite{fung2018mitigating} integrate anomaly detection mechanisms into the federated learning process. These methods analyze model updates received from clients to identify outliers that significantly deviate from the expected behavior.

AIJack allows Users to easily integrate anomaly detection methods like FoolsGold with FL protocols, as shown in Code~\ref{code:foolsgold}.

\begin{figure}[!ht]
\begin{lstlisting}[language=Python, caption=FoolsGold, label=code:foolsgold]
from aijack.collaborative import FedAVGClient, FedAVGServer, FedAVGAPI
from aijack.defense.foolsgold import FoolsGoldServerManager

clients = [FedAVGClient(NN()), FedAVGClient(NN())]
           
manager = FoolsGoldServerManager()
FoolsGoldFedAVGServer = manager.attach(FedAVGServer)
server = FoolsGoldFedAVGServer(clients, NN())

api = FedAVGAPI(server, clients, criterion, optimizers, dataloaders)
api.run()
\end{lstlisting}
\end{figure}


\end{document}